\documentclass{ifacconf}

\usepackage{graphicx}      
\usepackage{natbib}        
\usepackage{amsmath,mathtools}
\usepackage{amsfonts}
\usepackage{color}
\usepackage{multirow}
\begin{document}
\begin{frontmatter}

\title{Data-driven Driver Model for Speed Advisory Systems in Partially Automated Vehicles} 


\author[CAR]{Olivia Jacome,} 
\author[CAR]{Shobhit Gupta,} 
\author[CAR]{Stephanie Stockar,}
\author[CAR]{Marcello Canova}

\address[CAR]{Center for Automotive Research, The Ohio State University,  
   Columbus, OH 43212 USA (e-mail: jacome.2@osu.edu, gupta.852@osu.edu, stockar.1@osu.edu, canova.1@osu.edu).}
   
\begin{abstract}                
Vehicle control algorithms exploiting connectivity and automation, such as Connected and Automated Vehicles (CAVs) or Advanced Driver Assistance Systems (ADAS), have the opportunity to improve energy savings. However, lower levels of automation involve a human-machine interaction stage, where the presence of a human driver affects the performance of the control algorithm in closed loop. This occurs for instance in the case of Eco-Driving control algorithms implemented as a velocity advisory system, where the driver is displayed an optimal speed trajectory to follow to reduce energy consumption. Achieving the control objectives relies on the human driver perfectly following the recommended speed. If the driver is unable to follow the recommended speed, a decline in energy savings and poor vehicle performance may occur. This warrants the creation of methods to model and forecast the response of a human driver when operating in the loop with a speed advisory system.

This work focuses on developing a sequence to sequence long-short term memory (LSTM)-based driver behavior model that models the interaction of the human driver to a suggested desired vehicle speed trajectory in real-world conditions. A driving simulator is used for data collection and training the driver model, which is then compared against the driving data and a deterministic model. Results show close proximity of the LSTM-based model with the driving data, demonstrating that the model can be adopted as a tool to design human-centered speed advisory systems.

\end{abstract}


\end{frontmatter}
\section{Introduction} \label{sec1:intro}
The growth of technologies such as 5G communication networks, cloud computing, and artificial intelligence (AI), coupled with sensor fusion from global positioning systems (GPS) and cameras, has laid the foundation for the development of CAVs and ADAS. Despite these advancements, technical and liability challenges have slowed the mass deployment of fully automated vehicles in the transportation sector until 2030 \cite{article4}.  Humans therefore play a crucial role in facilitating the adoption of current ADAS systems, as they are becoming a major safety requirement. As the future of transportation relies more and more on the coupled behavior of humans and semi-autonomous vehicles, ADAS systems must be designed to be 'smart and predictable participants' that ensure passenger and surrounding vehicle safety. This applies not only to the vehicle control algorithms, but also to the human machine interface (HMI) as it is responsible for relaying cues to the driver. 
 
Existing work such as by \cite{manzoni2010driving} has cited driving behavior as one of the most important factors influencing fuel consumption by as much as 40\%. In the current generation of ADAS, as illustrated in \cite{malikopoulos2013optimization,fleming2018driver,linda2012improving}, drivers are displayed audio/video cues via HMIs, such as speed advisory systems that encourage a conservative driving style to improve energy savings. Several authors, including \cite{rojdestvenskiy2018real, wijayasekara2014driving}, have performed velocity trajectory optimization offline and have fed the optimal velocity to a speed advisory system for a driver to follow. The aforementioned methods were shown to improve the energy efficiency but were found to delay the arrival time at destination, and sacrifice the overall drivability \cite{malikopoulos2013optimization,wan2016optimal}. The major limitation of the existing work is that they do not predict or model the interaction of the human driver with the ADAS system  when part of a feedback control loop,  such as the case of speed advisory systems integrated with trajectory optimization. This leads to a mismatch in the velocity response (conservative vs. aggressive) prescribed by the speed advisory controller and the actual driver behavior, which ultimately leads to a decline in the potential energy savings and overall poor experience by the human driver. This highlights the necessity of developing methods that can learn the human driver response and utilize such information for the synthesis of the reference velocity trajectory provided to the driver.
A challenge with developing such models is understanding and mimicking the human driver behavior in changing traffic conditions, such as in the presence of speed limits, traffic lights, etc. In early work by \cite{cheung2018identifying}, the driver behavior was typically correlated with various background traits (e.g. age, questionnaire responses, gender, years of driving experience, violation records). Attempts have been made to develop deterministic models describing the response of a human driver within the context of specific driving scenarios. Noteworthy examples are the Gipps Model, the Intelligent Driver Model (IDM), and the Enhanced Driver Model (EDM) as illustrated in \cite{gipps1986model,gupta2019enhanced}. Data driven methods, such as recurrent neural networks (RNN), gated recurrent unit (GRU), and LSTM models have utilized vehicle features such as speed, acceleration, and headway to develop velocity prediction algorithms \cite{jiang2019trajectory,yunpeng2017multi,abi2021trip}. For example, \cite{qi2018connected} attempts at modeling driver error using a Markov Chain in the context of a stochastic model predictive controller (SMPC) for Eco-Driving, but this approach disregards the impact of the surrounding environment and traffic rules. Clustering techniques have also been used to identify driver error, but these techniques generalize driver behavior and do not predict individualized speed advisories \cite{wang2020driver}. An alternative data driven model in \cite{wijayasekara2015data} is developed by reshaping the optimal velocity advisory from historical fuel efficient data. This method however, is limited in application to the route studied.  

\begin{figure}[t!]
	\begin{center}
		\includegraphics[width=\columnwidth]{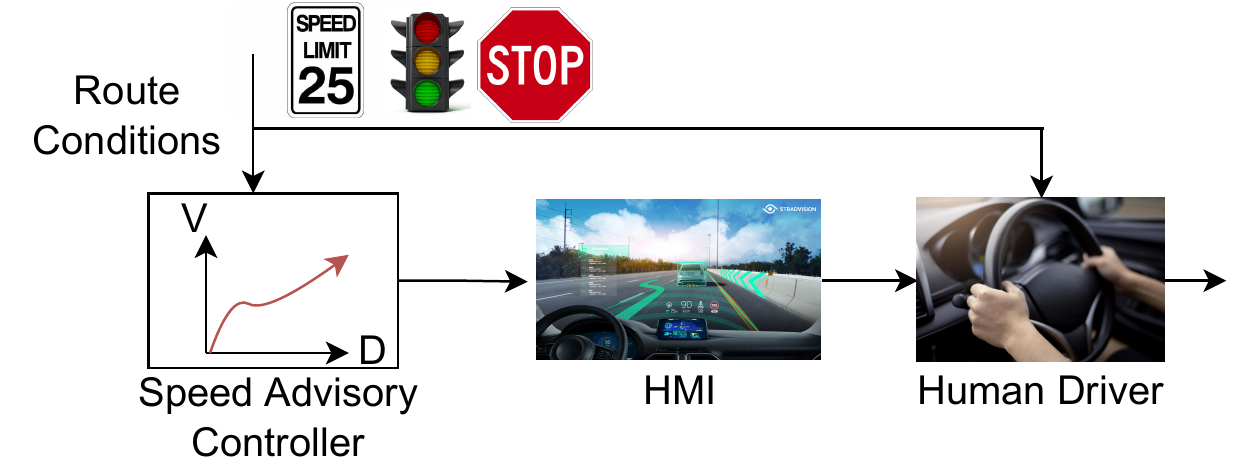}
		\caption{Driver Advisory System With Human in Open Loop}
		\label{fig:Hil_open}
	\end{center}
\end{figure}

 However, the human driver not only makes decision based on route information but also on the basis of speed recommendations as illustrated in Fig. \ref{fig:Hil_open}. To this end, it is noteworthy that none of the aforementioned models have been successful in capturing a human driver holistically, and adapting to the individual driver behavior under a speed advisory system, see \cite{abuali2016driver}.
 
This paper proposes a data-driven LSTM encoder-decoder (LSTMED) neural network driver behavior model that captures the driver's response when in the loop with a speed advisory system and in the presence of changing route conditions.  A comprarative study is performed where a deterministic approach based on an established Enhanced Driver Model is compared against the LSTMED driver behavior model. The LSTMED model is trained and tested using experimental data collected on a driving simulator. Both models are trained to predict the future velocity trajectory and the reference tracking error that the driver is likely to produce in the future. The proposed work will help the existing ADAS systems to learn, adapt and communicate effectively to different drivers, ultimately improving individual driver comfort, traffic flow and energy savings. As, the future of vehicle autonomy relies on the collaboration of humans and vehicle systems; the developed architecture can be applied to optimization strategies incorporating driver behavior and may be used as a framework to model the interaction of humans and systems as new technologies are developed.

\section{Driving Simulator and Data Collection} \label{sec2:data_collection}
To predict the response of a human driver to a speed advisory system, it is crucial to collect the driving data where the driver interacts with a HMI system. To ensure participant safety, data was collected in a controlled setting using the driver-in-loop (DiL) simulator at the Ohio State University Center For Automotive Research (OSU-CAR). 

The DiL simulator is a low latency, medium fidelity driving simulator that provides partial physical immersion via a 2018 Ford Mustang instrument panel cluster and A-pillars. Platform motion is controlled by 4 D-box gen two 4250i motion actuators providing 3 DoF in the roll, pitch and heave axes. The actuators provide a maximum of 3\textdegree  and 7\textdegree  of pitch and roll, respectively, while heave is provided up to +/-3inches. Three monitors centered at the drivers eye point provide a horizontal field of view of 9\textdegree and 12.6\textdegree view in the vertical direction. The driving scenarios are designed and rendered with SCANeR Studio, and the graphics render at refresh rates up to 120Hz. AV Simulation's SCANeR Studio is a comprehensive driving simulation software package developed for vehicle dynamics and road traffic research and development. The vehicle is modeled with CarSim and communicates with SCANeR via ComUDP at 1kHz providing the longitudinal and lateral dynamics of a 2020 Chevrolet Blazer. The DiL was validated in \cite{sekar2022assessment}, by comparing metrics calculated from the simulated longitudinal and lateral vehicle dynamics to corresponding data collected during on-road testing. The simulated terrain used for collection replicated the real world route up to 10m accuracy and was used to demonstrate longitudinal absolute and relative validity of the simulator. The verification conducted indicates that the simulator is a robust tool for data collection and assessment in a controlled and repeatable environment.   

\subsection{Terrain Creation}
\begin{figure}[htb]
	\begin{center}
		\includegraphics[width=\columnwidth]{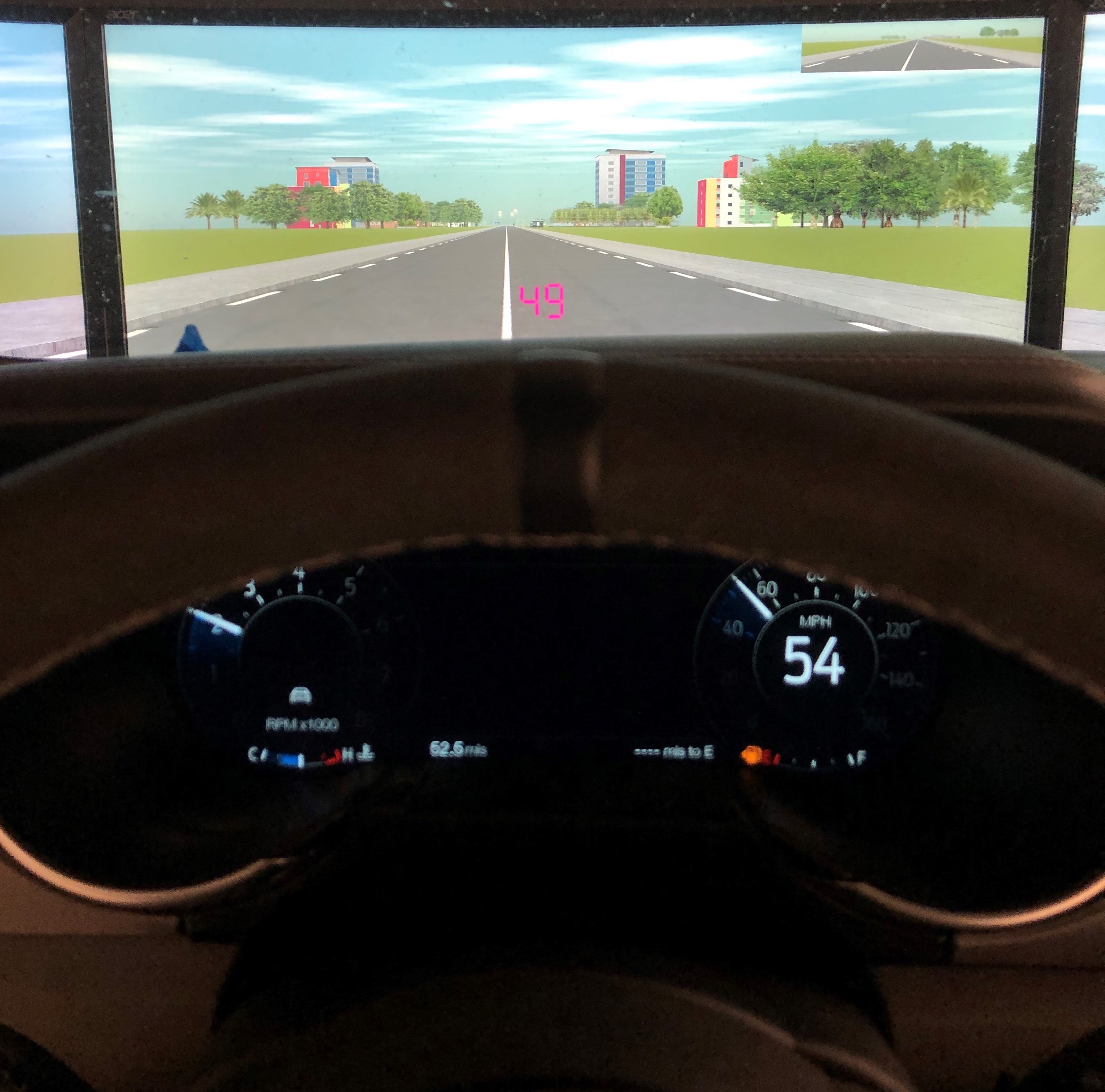}
		\caption{Simulator Driving view for Developed Terrain}
		\label{fig:sim_view}
	\end{center}
\end{figure}

To collect driving data, a 5 mile route consisting of urban and highway regions was developed in SCANeR Studio's Terrain Mode. The route consisted of 5 stops and contained speed limit regions that varied from 25mph to 50mph as seen in Fig. \ref{fig:sample data}. The straight line route contained regions of rural and city driving and took participants about 8 minutes to drive. The speed advisory was provided in red on the center screen above the steering wheel, as seen in Fig. \ref{fig:sim_view}. 

\subsection{Data Collection}
Thirty-six participants volunteered to interact with the driving simulator and generate data. In an effort to analyze the behavior of a human driver in the presence of a velocity advisory system, a reference profile was provided in real time to each driver to follow. The reference velocity profile was generated for the 5 mile route using the EDM, whose mathematical form is summarized below and detailed in \cite{gupta2019enhanced}. To introduce an additional source of variability in the test results, ten  calibrations of the EDM were applied to generate different reference velocity profiles that were randomly assigned to the drivers.

The participants were given 5 minutes to familiarize with the simulator in a default SCANeR city. Afterwards, the participants were asked to drive the route following one of the 10 reference profiles as best as they could, while obeying all other standard traffic laws. After a short break, each participant was asked to drive the route again with a new reference profile. After removing outliers, 71 total data sets were compiled. In the data collection phase, participants found difficulties to control the deceleration rate due to the stiffness of the brake pedal, and were therefore not likely to come to a complete stop at intersections. 

  \begin{figure}[t!]
	\begin{center}
		\includegraphics[width=\columnwidth]{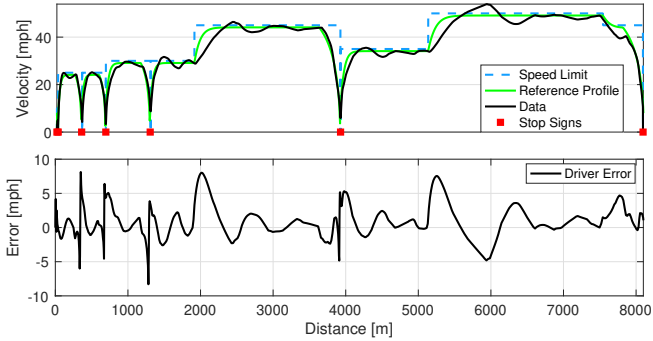}
		\caption{Sample Velocity and Error Profile for Simulator Driver}
		\label{fig:sample data}
	\end{center}
\end{figure}

Fig. \ref{fig:sample data} shows the velocity profile for one sample driver and the error with respect to the given speed advisory. An initial analysis of the data shows similar distribution of velocity values consistent with data collected in road vehicle studies, for example those presented in \cite{du2020vehicle}. 

\begin{figure}[t!]
	\begin{center}
		\includegraphics[width=\columnwidth]{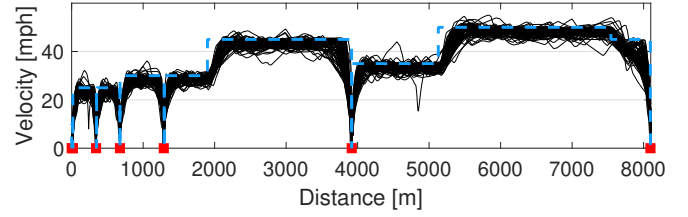}
		\caption{71 Driver Velocity Profiles}
		\label{fig:71_driver_pro}
	\end{center}
\end{figure}

All the 71 velocity traces collected on the driving simulator together with the route speed limit are shown in Fig. \ref{fig:71_driver_pro}. The data collected on the DiL simulator show generally less variability with respect to the tendency to overshoot the suggested velocity or speed limits, compared to the data collection studies without a speed advisory. This is likely due to the fact that participants were requested to focus on following the provided speed advisory during the tests.

\section{Driver Behavior Modeling for Velocity Advisory Systems} \label{sec3:driver_models}
As indicated in the DiL simulator study, human drivers tend to deviate from the suggested velocity due to the individual differences in perceiving the environment and speed advisory output, the aggressiveness in commanding the accelerator and brake pedal. For the development of predictive controls for human-in-loop ADAS algorithms, it is important to model the response of a human driver to a speed advisory or reference. This section compares a deterministic (EDM) and stochastic modeling approach, the latter based upon a LSTMED architecture.

\subsection{Deterministic Model: Enhanced Driver Model}
The EDM is a deterministic velocity predictor that can represent different rates of driver response, based on a set of parameters that can be calibrated from experimental data. The model, whose assumptions and derivation are elaborated in \cite{gupta2019enhanced}, includes three distinct operating modes (acceleration, deceleration and stop):

\begin{equation}\label{eq:edm_1}
\dfrac{dv_d(t)}{dt} = \left\{\begin{matrix}
a\left[1-\left(\frac{v_d(t)}{v_r(t)-\theta_0}\right)^\delta\right], & v_d(t)<v_r(t)\\
-b\left[1-\left(\frac{v_r(t)-\theta_0}{v_d(t)}\right)^\delta\right], & v_d(t)\geq v_r(t) \\
-\frac{1}{b}\left(\frac{v_d(t)^2}{2s(t)}\right)^2,  & s(t)<s_{brake} &
\end{matrix}\right.
\end{equation}
where $v_d(t)$ is the driver's velocity in response to a reference velocity value $v_r(t)$, which could be assumed as the route speed limits or the input provided by the speed advisory. The parameters $a$, $b$, $\delta$ and $\theta_0$ determine the response to the input reference and represent the maximum acceleration, deceleration, aggressiveness and the offset from the reference speed, respectively. 

Equations \label{eq:edm_1} represent the driver behavior in freeway driving mode, namely with absence of traffic. In presence of a lead vehicle, the EDM equations change slightly to model a car-following mode scenario, where the reference speed is assumed to be the lead vehicle velocity. The last operating mode (stop mode) ensures the vehicle decelerates safely a few meters before a stop. The variable $s(t)$ in the stop mode equation is the distance of the vehicle from the next stop on its route. The EDM enters this mode when $s(t)$ is lower than the critical braking distance $s_{brake}$, as defined in \cite{gupta2019enhanced}.

The EDM has been validated with experimental data collected on instrumented vehicles, demonstrating its capability of representing real-world driver data, as detailed in \cite{gupta2020estimation,rajakumar2020benchmarking}. 

\subsection{Stochastic Model: LSTM Encoder-Decoder Model}
Existing work on modeling driver behavior involves developing a model able to predict common patterns in driving maneuvers. Models of this kind can be used to detect driving patterns for a typical driver, but do not focus on modeling the behavior of the driver while interacting with an HMI such as a speed advisory system. The short term planning of a typical human driver with respect to traffic conditions is generally represented as a time-varying, nonlinear time-series in terms of speed and acceleration \cite{gupta2021eco}. In this work, the driver behavior in response to a speed advisory is modeled as a sequence learning task using LSTM Encoder-Decoder architecture. 

\subsubsection{LSTMED Architecture:} LSTM is a technique proposed by \cite{hochreiter1997long} as a response to the inability of conventional RNNs' to efficiently handle large time span dependencies due to the problem of vanishing gradients. The basic structure of LSTM unit, as illustrated in \cite{hochreiter1997long}, comprises of an input gate, output gate, forget gate, and a memory cell that selectively controls the flow of information to produce a hidden state. LSTM units are stacked as per the input sequence length and number of units to produce a stacked LSTM layer. The LSTMED is a neural network architecture that consist of two LSTM layers, i.e., the encoder LSTM layer ($\text{N}_{\text{E}}$ units) and the decoder LSTM layer ($\text{N}_{\text{D}}$ units). It first encodes the input sequence comprising of historical sequence of $t_h$ seconds ($N_{\text{H}}$ points) and $\text{N}_{\theta,x}$ input features,  $X=(x_{t-t_h},\cdots,x_{t}) \in \mathbb{R}^{\text{N}_{\theta,x}\times \text{N}_{\text{H}}}$ into a fixed-length hidden vector representation $h_t\in\mathbb{R}^{\text{N}_\text{H}\times \text{N}_\text{E}}$. This is then fed to a dropout layer which excludes $20\%$ of the neurons to avoid over fitting. Finally, the hidden state is passed to the decoder layer, after which a linear transformation is performed to predict the output sequence of $t_p$ seconds ($\text{N}_\text{P}$ points) and $\text{N}_{\theta,y}$ output features, $Y=(y_t\cdots y_{t+t_p})\in \mathbb{R}^{\text{N}_{\theta,y}\times \text{N}_\text{P}}$. 

\begin{figure}[t!]
	\begin{center}
		\includegraphics[width=\columnwidth]{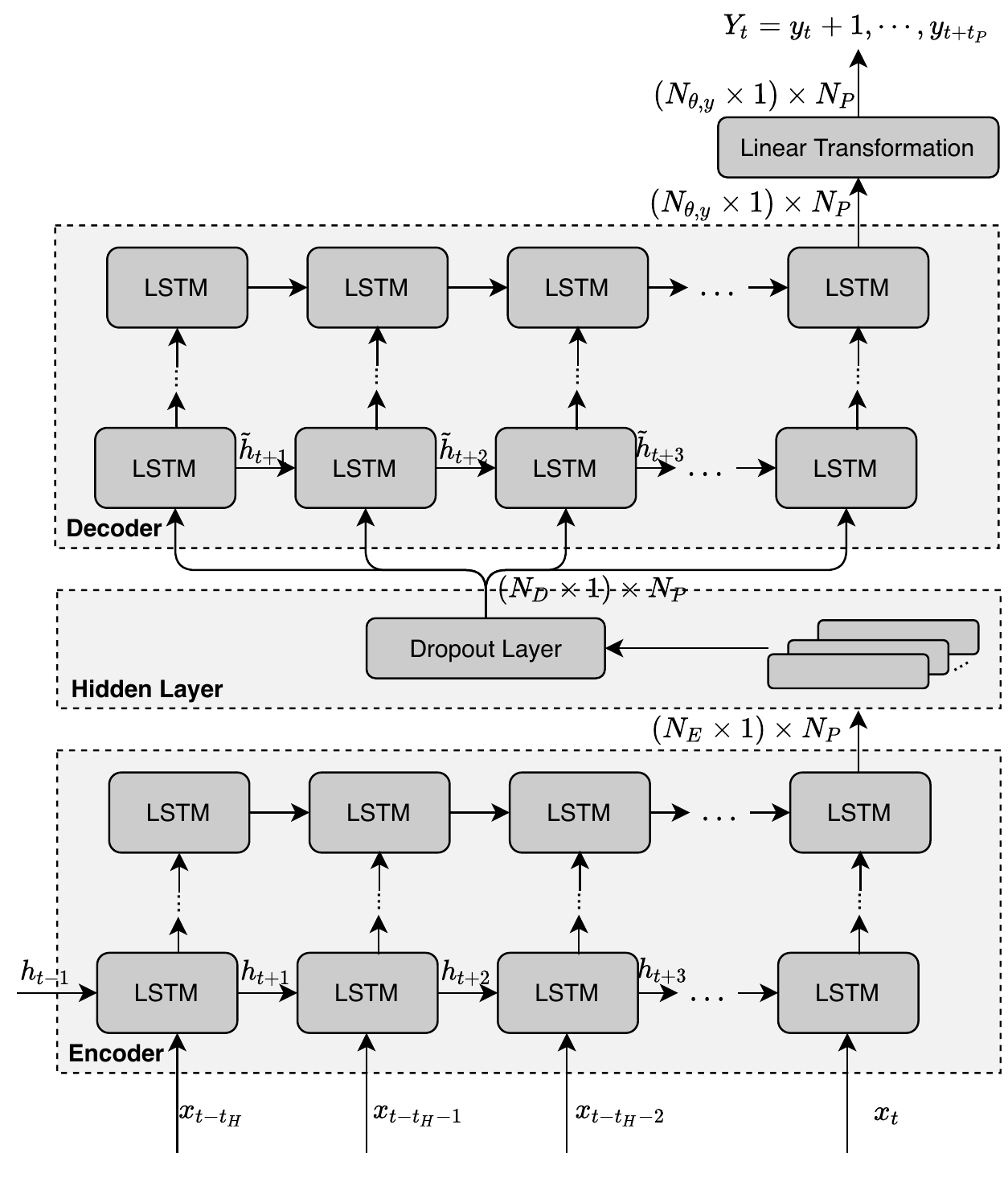}
		\caption{LSTM Encoder Decoder Driver Model}
		\label{fig:lstmed}
	\end{center}
\end{figure}

The given architecture is unique from typical LSTM prediction models since it takes an iterative approach in sequence prediction. In this application, the LSTMED predictions are unrolled based on the last $t-1$ prediction step in an iterative manner. In developing the LSTMED, different architectures were evaluated before settling on the structure shown in Fig. \ref{fig:lstmed}, which produced the smallest RMSE for both velocity and error prediction.

\subsubsection{Input and Output Features:}
The crucial step in developing the LSTMED driver model is the selection of the input features that can represent real-world driving with recommendations from a speed advisory system. In this paper, input features that can be easily measured over CAN bus have been selected. Moreover, it is assumed that the ego vehicle is connected and can receive SPaT messages from signalized intersections. The longitudinal motion of any vehicle on a straight and flat road over the next time step $\Delta t$ can be determined by velocity $v_t$ and acceleration $a_t$ at current time $t$:

\begin{equation}
	\label{eqn: longitudinal_dynamics}
	\begin{aligned}
		v_{t+1}=v_t+a_t\Delta t.
	\end{aligned}
\end{equation}

Hence, $[v_t, a_t]$ can be used to develop non-linear models representing human-driving which have also been shown in majority of car-following longitudinal models (\cite{gupta2019enhanced, rajakumar2020benchmarking}). Further, distance to the traffic light $d_{TL,t}^e$ and signal phase $\tau_{SP}$ is considered to comprehensively represent human driving on a straight road and near signalized intersections. Velocity reference $v_t^{ref}$ and tracking error ($\delta_t^v$) between the reference and vehicle speed is assumed to the two additional input features that represent the interaction of the human driver with the speed advisory system. Vehicle velocity and tracking error are the two features that are predicted by the LSTMED model to capture this interaction. Finally, the input feature set,  $x_t = \{v_t, a_t, d_{TL,t}^e, v_t^{ref},\tau_{SP}, \delta_t^v\}$ can comprehensively represent human driving while interacting with a speed advisory system, and $y_t = \{v_t, \delta_t^v\}$ represent the output feature set predicted by LSTMED model.

\section{Simulation Results and Analysis}
\subsection{Training Methodology}
\begin{figure}[t!]
	\begin{center}
		\includegraphics[width=\columnwidth]{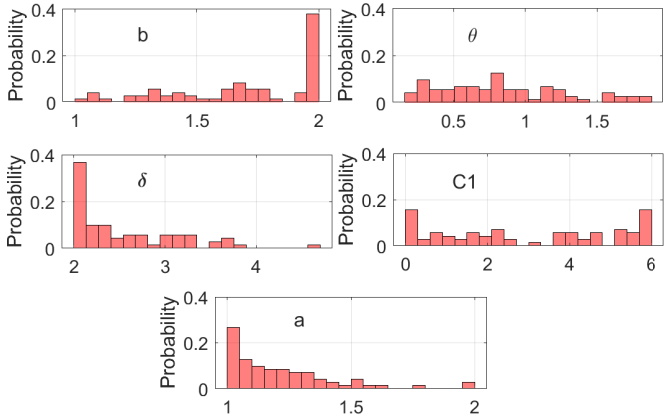}
		\caption{Calibrated Distribution of EDM Parameters}
		\label{fig:EDM_dist}
	\end{center}
\end{figure}

The LSTMED model was trained over 66 datasets and the remaining 5 datasets were randomly chosen for testing using the PyTorch package, illustrated in \cite{paszke2019pytorch}. For better generalization over the different features, the feature set $x_t$ is normalized between 0 and 1 using sci-kit learn toolbox in Python. Custom functions were designed to split the data into a historical window and prediction window of time duration $t_H$ and $t_P$ respectively during the training phase. $t_H$ and $t_P$ were varied as in \cite{gupta2021eco} to find the ideal time series windows of 30 s and 5 s respectively. To evaluate the performance of the prediction model, Root Mean Square Error (RMSE) is computed between the predicted speed ($\hat y_t$) and the actual speed ($ y_t$):
\begin{equation}
	\label{eqn: RMSE}
	\text{RMSE} = \sqrt{\frac{1}{N}\sum_{i=1}^{N}(y_t-\hat y_t)^2}
\end{equation}
To perform the comparison of LSTMED with the EDM, the latter was calibrated for all 71 driver profiles collected in section \ref{sec2:data_collection} using a Genetic Algorithm as explained in \cite{gupta2020estimation}. The distribution of the calibrated parameters for all collected data sets have been summarized in Fig. \ref{fig:EDM_dist}. The majority of the drivers were on the more relaxed side of acceleration and more aggressive for deceleration. No participants had an offset greater than 2 $m/s$. Lower values for the parameter $\delta$ reveal that participants were less aggressive during accelerations, while lower value of $C1$ indicate smaller braking distance and a more aggressive behavior while braking. The results reveal that the drivers were aggressive on the brake pedals while approaching stops, which may be due to the issues experienced with the brake pedal on the DiL simulator. The distribution of parameters and values aligns with those observed in \cite{gupta2020estimation}, which further validates the calibration process.

\subsection{Comparison of Driver Behavior Models}
A comparison was performed against the experimental data and prediction results from the EDM. For conciseness, three drivers were chosen from the 5 test cases for illustrating the results. The typical route speed limit of the calibrated EDM was replaced with the reference profile as portrayed to the selected drivers during simulator data collection via the HMI. This allowed the EDM driver error to be collected as the difference in the velocity projected by the EDM and the displayed reference. Historical features collected from the simulated run for the selected drivers were simultaneously fed into the LSTMED model for driver velocity and error predictions. 

\begin{figure}[t!]
	\begin{center}
		\includegraphics[width=\columnwidth]{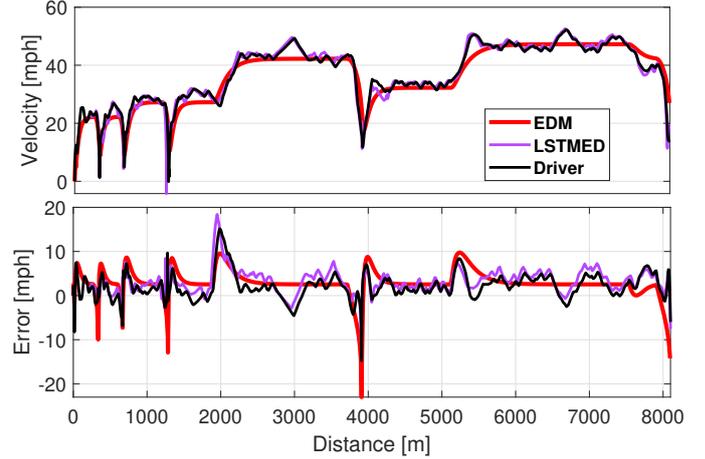}
		\caption{Sample EDM and LSTM Predictor Results}
		\label{fig:EDM_LSTM_predicitons_sample}
	\end{center}
\end{figure}

Fig. \ref{fig:EDM_LSTM_predicitons_sample} displays the point to point EDM and LSTMED prediction results compared to the actual driver 3 data for velocity and driver error with respect to a speed advisory. 
It can be seen that the LSTMED predictions are in closer proximity to the driving data, where it captures the oscillations and overshoots of the human driver as compared to the EDM model. 

\begin{figure}[t!]
	\begin{center}
		\includegraphics[width=\columnwidth]{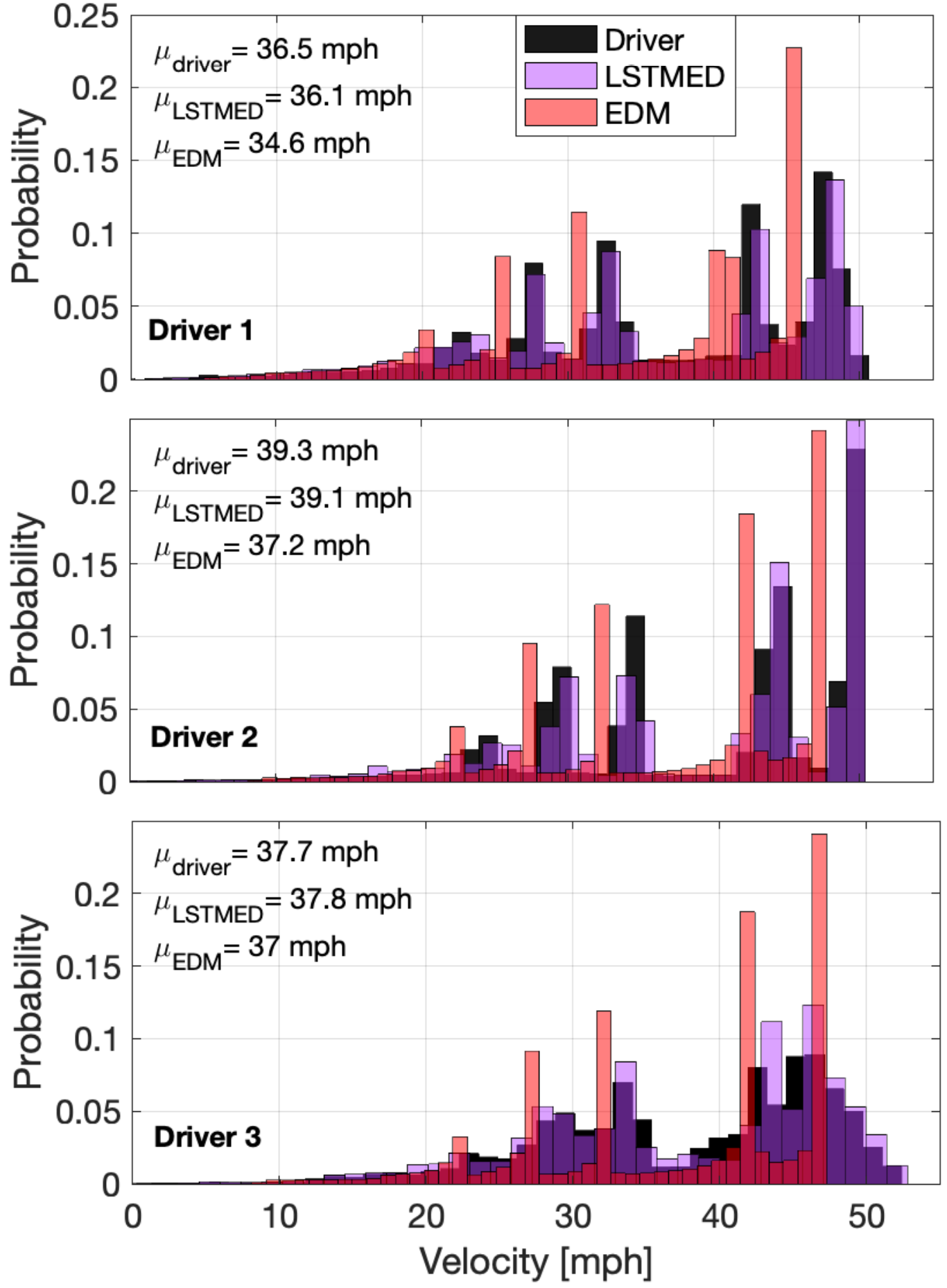}
		\caption{LSTMED and EDM Velocity Predictions vs Actual Simulator Driving Data}
		\label{fig:vel_dists_3_drivers}
	\end{center}
\end{figure}

\begin{figure}[t!]
	\begin{center}
		\includegraphics[width=\columnwidth]{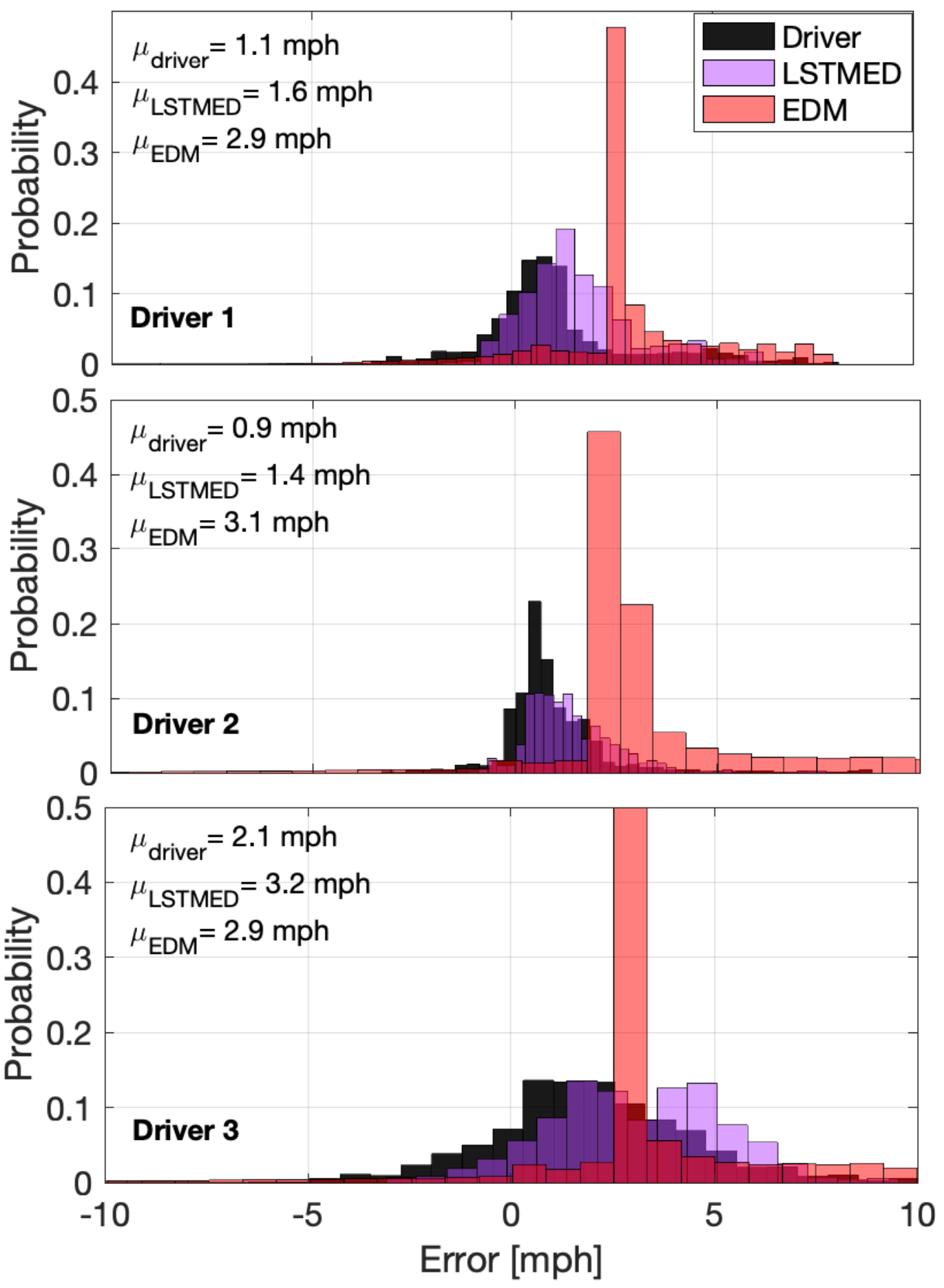}
		\caption{LSTMED and EDM Driver Error Predictions vs Actual Simulator Driving Data}
		\label{fig:error_dists_3_drivers}
	\end{center}
\end{figure}

The distribution of the actual driver, LSTMED and EDM errors for velocity and driver error with respect to the same reference can be seen in \ref{fig:error_dists_3_drivers} and \ref{fig:vel_dists_3_drivers}. The mean error and velocity for the LSTMED prediction was observed to be closer to the driving data while the EDM mean error was very different. In addition, the overall velocity and error distribution modes for LSTMED were similar to the driving data. The EDM model did not capture this variation and typically has a higher standard deviation compared to the actual data for each of the three test drivers selected.

Furthermore, the histograms indicate that the distribution of the reference tracking error for all drivers predicted by the LSTMED is similar to the driving data. Table \ref{tab:RMSE Vel 3 drivers} summarizes the RMSE of the EDM and LSTMED velocity and reference tracking error predictions with respect to the actual data. As the EDM error is derived from the velocity, RMSE values for both velocity and error prediction are similar. Overall the LSTMED predictions for reference tracking error and velocity are minimal compared to the EDM where the second driver's velocity prediction can be considered an outlier. Overall, the results signify that the LSTMED has the potential to predict the response of a human driver to a speed advisory system, with minimal prediction error from real-world setting.

\begin{table}[t]
\centering
\caption{RMSE LSTM and EDM Driver Velocity and Error Prediction}	
\label{tab:RMSE Vel 3 drivers}
\begin{tabular}{|l|ll|ll|}
\hline
\multirow{2}{*}{Driver ID} & \multicolumn{2}{l|}{Velocity}    & \multicolumn{2}{l|}{Velocity Error} \\ \cline{2-5} 
                           & \multicolumn{1}{l|}{LSTM} & EDM  & \multicolumn{1}{l|}{LSTM}   & EDM   \\ \hline
Driver 1                   & \multicolumn{1}{l|}{2.46} & 2.84 & \multicolumn{1}{l|}{1.61}   & 2.84  \\ \hline
Driver 2                   & \multicolumn{1}{l|}{3.35} & 3.19 & \multicolumn{1}{l|}{1.43}   & 3.19  \\ \hline
Driver 3                   & \multicolumn{1}{l|}{1.72} & 3.04 & \multicolumn{1}{l|}{1.93}   & 3.04  \\ \hline
\end{tabular}
\end{table}

\section{Conclusion}
In this paper, a data-driven sequence-to-sequence LSTMED driver model was developed to model the interaction of a real-world driver to human machine interfaces, specifically a speed advisory algorithm in the context of ADAS. The developed model was trained using the real-world driving data collected on a Driver in the Loop simulator.

The trained model predicts the velocity and the reference tracking error in close proximity to the real-world driver and outperformed a deterministic model (EDM) commonly utilized in literature. 

This work is an initial step towards the development of predictive control strategies for ADAS systems catered towards human in the loop applications. Specifically, the model would help in adapting a speed advisory system to better match the perception and response of the human by considering the predicted driver error and velocity with respect to the advisory reference.  Future work includes expanding the data set and incorporating features from the surrounding  vehicles and traffic in the environment. 


\begin{ack}
The authors acknowledge the support from the Advanced Research Projects Agency – Energy (ARPA-E) NEXTCAR project (Award Number DE-AR0000794). The authors are grateful to BorgWarner Inc., for the helpful discussions during this research.\end{ack}

\bibliography{ifacconf.bib}             

\begin{thebibliography}{25}
\providecommand{\natexlab}[1]{#1}
\providecommand{\url}[1]{\texttt{#1}}
\providecommand{\urlprefix}{URL }
\expandafter\ifx\csname urlstyle\endcsname\relax
  \providecommand{\doi}[1]{doi:\discretionary{}{}{}#1}\else
  \providecommand{\doi}{doi:\discretionary{}{}{}\begingroup
  \urlstyle{rm}\Url}\fi

\bibitem[{Abi~Akl et~al.(2021)Abi~Akl, El~Khoury, and Mansour}]{abi2021trip}
Abi~Akl, N., El~Khoury, J., and Mansour, C. (2021).
\newblock Trip-based prediction of hybrid electric vehicles velocity using
  artificial neural networks.
\newblock In \emph{2021 IEEE 3rd International Multidisciplinary Conference on
  Engineering Technology (IMCET)}, 60--65. IEEE.

\bibitem[{AbuAli and Abou-zeid(2016)}]{abuali2016driver}
AbuAli, N. and Abou-zeid, H. (2016).
\newblock Driver behavior modeling: Developments and future directions.
\newblock \emph{International journal of vehicular technology}, 2016.

\bibitem[{Cheung et~al.(2018)Cheung, Bera, Kubin, Gray, and
  Manocha}]{cheung2018identifying}
Cheung, E., Bera, A., Kubin, E., Gray, K., and Manocha, D. (2018).
\newblock Identifying driver behaviors using trajectory features for vehicle
  navigation.
\newblock In \emph{2018 IEEE/RSJ International Conference on Intelligent Robots
  and Systems (IROS)}, 3445--3452. IEEE.

\bibitem[{Du et~al.(2020)Du, Cui, Li, Nie, Shi, Wang, and Li}]{du2020vehicle}
Du, Y., Cui, N., Li, H., Nie, H., Shi, Y., Wang, M., and Li, T. (2020).
\newblock The vehicle’s velocity prediction methods based on rnn and lstm
  neural network.
\newblock In \emph{2020 Chinese Control And Decision Conference (CCDC)},
  99--102. IEEE.

\bibitem[{Fleming et~al.(2018)Fleming, Yan, Allison, Stanton, and
  Lot}]{fleming2018driver}
Fleming, J., Yan, X., Allison, C., Stanton, N., and Lot, R. (2018).
\newblock Driver modeling and implementation of a fuel-saving adas.
\newblock In \emph{2018 IEEE International Conference on Systems, Man, and
  Cybernetics (SMC)}, 1233--1238. IEEE.

\bibitem[{Gipps(1986)}]{gipps1986model}
Gipps, P.G. (1986).
\newblock A model for the structure of lane-changing decisions.
\newblock \emph{Transportation Research Part B: Methodological}, 20(5),
  403--414.

\bibitem[{Gupta and Canova(2021)}]{gupta2021eco}
Gupta, S. and Canova, M. (2021).
\newblock Eco-driving of connected and autonomous vehicles with
  sequence-to-sequence prediction of target vehicle velocity.
\newblock \emph{IFAC-PapersOnLine}, 54(10), 430--436.

\bibitem[{Gupta et~al.(2019)Gupta, Deshpande, Tulpule, Canova, and
  Rizzoni}]{gupta2019enhanced}
Gupta, S., Deshpande, S.R., Tulpule, P., Canova, M., and Rizzoni, G. (2019).
\newblock An enhanced driver model for evaluating fuel economy on real-world
  routes.
\newblock \emph{IFAC-PapersOnLine}, 52(5), 574--579.

\bibitem[{Gupta et~al.(2020)Gupta, Deshpande, Tufano, Canova, Rizzoni, Aggoune,
  Olin, and Kirwan}]{gupta2020estimation}
Gupta, S., Deshpande, S.R., Tufano, D., Canova, M., Rizzoni, G., Aggoune, K.,
  Olin, P., and Kirwan, J. (2020).
\newblock Estimation of fuel economy on real-world routes for next-generation
  connected and automated hybrid powertrains.
\newblock Technical report, SAE Technical Paper.

\bibitem[{Hochreiter and Schmidhuber(1997)}]{hochreiter1997long}
Hochreiter, S. and Schmidhuber, J. (1997).
\newblock Long short-term memory.
\newblock \emph{Neural computation}, 9(8), 1735--1780.

\bibitem[{Intelligence(2022)}]{article4}
Intelligence, M. (2022).
\newblock Autonomous (driverless) car market - growth, trends, covid-19 impact,
  and forecast (2022 - 2027).

\bibitem[{Jiang et~al.(2019)Jiang, Chang, Li, and Chen}]{jiang2019trajectory}
Jiang, H., Chang, L., Li, Q., and Chen, D. (2019).
\newblock Trajectory prediction of vehicles based on deep learning.
\newblock In \emph{2019 4th International Conference on Intelligent
  Transportation Engineering (ICITE)}, 190--195. IEEE.

\bibitem[{Linda and Manic(2012)}]{linda2012improving}
Linda, O. and Manic, M. (2012).
\newblock Improving vehicle fleet fuel economy via learning fuel-efficient
  driving behaviors.
\newblock In \emph{2012 5th International Conference on Human System
  Interactions}, 137--143. IEEE.

\bibitem[{Malikopoulos and Aguilar(2013)}]{malikopoulos2013optimization}
Malikopoulos, A.A. and Aguilar, J.P. (2013).
\newblock An optimization framework for driver feedback systems.
\newblock \emph{IEEE Transactions on Intelligent Transportation Systems},
  14(2), 955--964.

\bibitem[{Manzoni et~al.(2010)Manzoni, Corti, De~Luca, and
  Savaresi}]{manzoni2010driving}
Manzoni, V., Corti, A., De~Luca, P., and Savaresi, S.M. (2010).
\newblock Driving style estimation via inertial measurements.
\newblock In \emph{13th International IEEE Conference on Intelligent
  Transportation Systems}, 777--782. IEEE.

\bibitem[{Paszke et~al.(2019)Paszke, Gross, Massa, Lerer, Bradbury, Chanan,
  Killeen, Lin, Gimelshein, Antiga et~al.}]{paszke2019pytorch}
Paszke, A., Gross, S., Massa, F., Lerer, A., Bradbury, J., Chanan, G., Killeen,
  T., Lin, Z., Gimelshein, N., Antiga, L., et~al. (2019).
\newblock Pytorch: An imperative style, high-performance deep learning library.
\newblock \emph{Advances in neural information processing systems}, 32.

\bibitem[{Qi et~al.(2018)Qi, Wang, Wu, Boriboonsomsin, and
  Barth}]{qi2018connected}
Qi, X., Wang, P., Wu, G., Boriboonsomsin, K., and Barth, M.J. (2018).
\newblock Connected cooperative ecodriving system considering human driver
  error.
\newblock \emph{IEEE Transactions on Intelligent Transportation Systems},
  19(8), 2721--2733.

\bibitem[{Rajakumar~Deshpande et~al.(2020)Rajakumar~Deshpande, Gupta, Kibalama,
  Pivaro, and Canova}]{rajakumar2020benchmarking}
Rajakumar~Deshpande, S., Gupta, S., Kibalama, D., Pivaro, N., and Canova, M.
  (2020).
\newblock Benchmarking fuel economy of connected and automated vehicles in real
  world driving conditions via monte carlo simulation.
\newblock In \emph{Dynamic Systems and Control Conference}, volume 84270,
  V001T10A004. American Society of Mechanical Engineers.

\bibitem[{Rojdestvenskiy et~al.(2018)Rojdestvenskiy, Cvetkovi{\'c}, and
  Bouchner}]{rojdestvenskiy2018real}
Rojdestvenskiy, D., Cvetkovi{\'c}, M., and Bouchner, P. (2018).
\newblock Real-time driver advisory system for improving energy economy based
  on advance driver assistant systems interface.
\newblock In \emph{2018 Smart City Symposium Prague (SCSP)}, 1--6. IEEE.

\bibitem[{Sekar et~al.(2022)Sekar, Jacome, Chrstos, and
  Stockar}]{sekar2022assessment}
Sekar, R., Jacome, O., Chrstos, J., and Stockar, S. (2022).
\newblock Assessment of driving simulator for longitudinal vehicle dynamics
  evaluation.
\newblock Technical report, SAE Technical Paper.

\bibitem[{Wan et~al.(2016)Wan, Vahidi, and Luckow}]{wan2016optimal}
Wan, N., Vahidi, A., and Luckow, A. (2016).
\newblock Optimal speed advisory for connected vehicles in arterial roads and
  the impact on mixed traffic.
\newblock \emph{Transportation Research Part C: Emerging Technologies}, 69,
  548--563.

\bibitem[{Wang et~al.(2020)Wang, Liao, Wang, Oswald, Wu, Boriboonsomsin, Barth,
  Han, Kim, and Tiwari}]{wang2020driver}
Wang, Z., Liao, X., Wang, C., Oswald, D., Wu, G., Boriboonsomsin, K., Barth,
  M.J., Han, K., Kim, B., and Tiwari, P. (2020).
\newblock Driver behavior modeling using game engine and real vehicle: A
  learning-based approach.
\newblock \emph{IEEE Transactions on Intelligent Vehicles}, 5(4), 738--749.

\bibitem[{Wijayasekara et~al.(2014)Wijayasekara, Manic, and
  Gertman}]{wijayasekara2014driving}
Wijayasekara, D., Manic, M., and Gertman, D. (2014).
\newblock Driving behavior prompting framework for improving fuel efficiency.
\newblock In \emph{2014 7th International Conference on Human System
  Interactions (HSI)}, 55--60. IEEE.

\bibitem[{Wijayasekara et~al.(2015)Wijayasekara, Manic, and
  Gertman}]{wijayasekara2015data}
Wijayasekara, D., Manic, M., and Gertman, D. (2015).
\newblock Data driven fuel efficient driving behavior feedback for fleet
  vehicles.
\newblock In \emph{2015 8th International Conference on Human System
  Interaction (HSI)}, 75--81. IEEE.

\bibitem[{Yunpeng et~al.(2017)Yunpeng, Di, Junpeng, and
  Yong}]{yunpeng2017multi}
Yunpeng, L., Di, H., Junpeng, B., and Yong, Q. (2017).
\newblock Multi-step ahead time series forecasting for different data patterns
  based on lstm recurrent neural network.
\newblock In \emph{2017 14th web information systems and applications
  conference (WISA)}, 305--310. IEEE.

\end{thebibliography}

\end{document}